\documentclass{article}
\usepackage{spconf}
\usepackage{amsmath}
\usepackage{graphicx}
\usepackage{epsfig}
\usepackage{epstopdf}
\usepackage{subfigure}
\usepackage{amssymb} 
\newcommand{\argmin}{\operatornamewithlimits{argmin}}
\newcommand{\argmax}{\operatornamewithlimits{argmax}}
\usepackage{algorithm}
\usepackage{algpseudocode}
\usepackage{color}
\usepackage{url}
\usepackage{balance}

\title{Action recognition in still images by latent superpixel classification}
\name{Shaukat Abidi, Massimo Piccardi, Mary-Anne Williams}
\address{Faculty of Engineering and IT, University of Technology, Sydney}
\begin{document}
%
\maketitle
\begin{abstract}
Action recognition from still images is an important task of computer vision applications such as image annotation, robotic navigation, video surveillance and several others. Existing approaches mainly rely on either bag-of-feature representations or articulated body-part models. However, the relationship between the action and the image segments is still substantially unexplored. For this reason, in this paper we propose to approach action recognition by leveraging an intermediate layer of ``superpixels'' whose latent classes can act as attributes of the action. In the proposed approach, the action class is predicted by a structural model(learnt by Latent Structural SVM) based on measurements from the image superpixels and their latent classes. Experimental results over the challenging Stanford 40 Actions dataset report a significant average accuracy of $74.06\%$ for the positive class and $88.50\%$ for the negative class, giving evidence to the performance of the proposed approach\footnote{ \textcopyright2015 IEEE}.
\end{abstract}

\section{Introduction and related work}
\label{sec:intro}
Automated recognition of actions in still images can play an important role for annotation of image catalogues, including the large collections of images which are increasingly made available by social networks. Actions which can be plausibly recognised from still imagery are those inferrable from the actors' poses and the presence of relevant objects: examples range from ``taking a picture'' and ``having a barbecue'', to ``throwing a javelin'' or ``playing guitar''. Moreover, recognition from single frames could also prove of fundamental value for recognising actions in video. For instance, in surveillance videos it is not uncommon to clearly sight an actor for only a few frames due to repeated occlusions. In such cases, it is not easy to recognise the action in dynamical terms, i.e., as the temporal evolution of a measurement vector. Rather, inference must be obtained as the cumulative evidence from a (possibly small) set of individual frames. Also in robotics, the varying camera viewpoint may make it easier to recognise actions from isolated frames than from sequences. Estimation from still frames is therefore a foundational technology for all these cases.

\begin{figure}[t]
\label{fig:gm}

\begin{minipage}[b]{1.0\linewidth}
  \centering
  \centerline{\label{a1}\includegraphics[width=8cm]{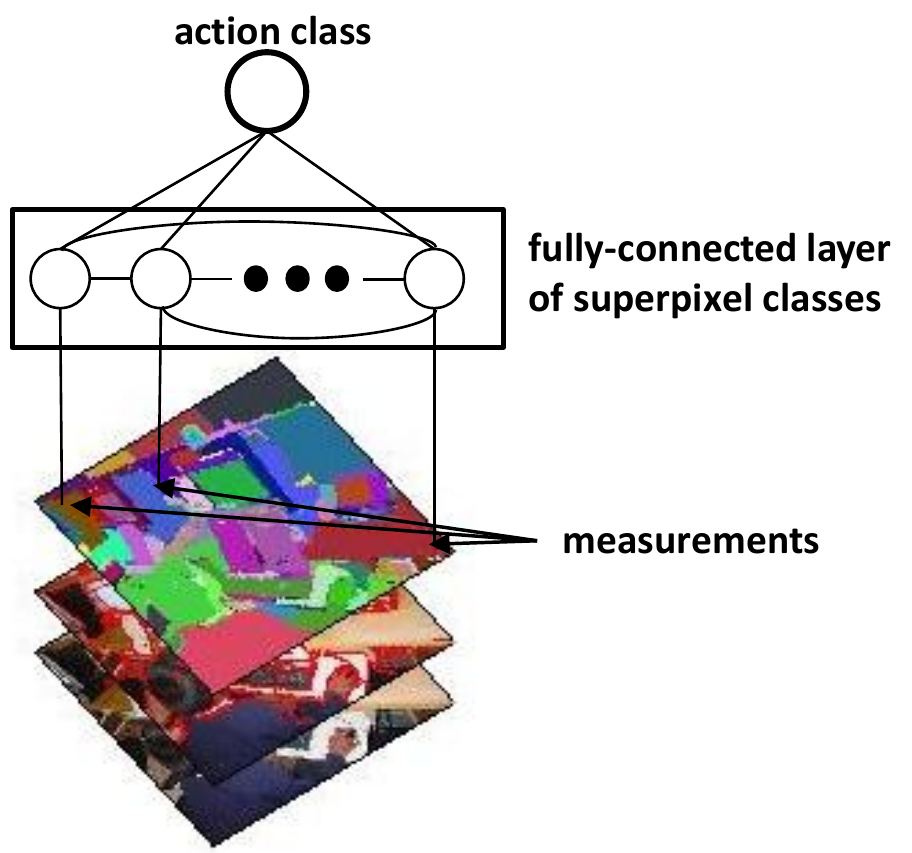}}
\end{minipage}
\caption{Action recognition: bottom layer: superpixel segmentation and feature extraction; intermediate layer: superpixel classification; top variable: action class.}
\end{figure}

The most straightforward solution to recognise actions from still images is to compute a bag-of-feature representation of the image and use it for classifying it into a relevant set of action classes~\cite{LaptevBMVC2010,IkizlerICPR2008,laptev2005space}. Useful features include local texture descriptors such as the histogram of oriented gradients (HOG), dense SIFT, GIST and several others~\cite{Dalal2005,lowe2004distinctive,Oliva2001}. Bag-of-features analysis usually discards the spatial coordinates at which descriptors are collected since it focusses on textural rather than spatial or structural information. These approaches have reported very interesting results on challenging still image action datasets such as those described in~\cite{IkizlerICCV2009,GuptaPAMI2009,LiFeiFeiCVPR2010}. At the opposite end of the spectrum are approaches based on the explicit recovery of body parts and the incorporation of structural information in the recognition process~\cite{MoriCVPR2010,LiFeiFeiICCV2011}. The baseline model is a latent part-based model akin to Pictorial Structure which can be estimated as a joint, conditional or max-margin model~\cite{PictorialStructures1973,PedroPAMI2010}. Delaitre \textit{et al}. has reported a comparison between a bag-of-features and a structural approach, showing that hybridisation of the two can be a way to capture the benefits of both models~\cite{LaptevBMVC2010}. A recent survey from Guo and Lai offers a comprehensive outline of the research in this area~\cite{Guo2014}.

Overall, it appears that the existing approaches have not substantially explored the underlying relationship between the action class and the segments of the containing image. For this reason, in this paper we propose to approach action recognition in still images by leveraging the latent classes of the image's ``superpixels'' (homogenous regions obtained from over-segmentation of the image~\cite{felzenszwalb2004efficient}). To this aim, we have designed a graphical model with the action as its root node and a fully-connected layer of superpixel classes to capture the relationship with the image segments~(Fig.~\ref{fig:gm}). A rich measurement vector is extracted from each superpixel, and model training is provided by a latent structural SVM approach.

The rest of the paper is organised as follows: Section 2 describes the graphical model for action recognition and the detectors for the superpixels' classes. Section 3 overviews the latent structural SVM framework. Section 4 describes the experiments and discusses results. Section 5 highlights conclusions and future work.

\section{Action recognition by superpixel classification}

The fundamental step taken in our approach is the decomposition of an image into small and coherent patches commonly referred to as superpixels. Our underlying assumption is that certain actions can be recognised effectively from an image by utilising useful information from the superpixels. Given their homogeneous nature, superpixels can be assigned single class labels of the type of ``sky'', ``road'', ``face'' and others, leading to a form of image (over-)segmentation~(Fig.~\ref{fig:gm}, bottom layers). While each superpixel in an image can be classified individually using a trained classifier, a recent paper from Pei \textit{et al}.~\cite{pei2013efficient} has shown that superpixel classification proves more accurate if all the superpixels in the image are classified jointly. Accordingly, our approach consists of two main stages: in the first stage, we pre-train a superpixel classifier, or a set of object detectors, from a supervised set of image regions and we use it to compute class scores for all superpixels in a given image. In the second stage, such scores are used as measurements in a graphical model that provides optimal, joint decisions for all superpixels and the action class. In the rest of this section, we describe the graphical model and the object detectors.



\subsection{The graphical model}
The proposed graphical model comprises three sets of variables, namely measurements ($x$), hidden nodes, or states, ($h$) and an output node ($y$). The measurements are a vector of detector scores for each superpixel, the hidden nodes are their classes, and the output node is the categorical variable for the action. The nodes are connected by three different types of edges: a) edges connecting measurements and states, b) edges over state pairs, and c) edges between states and the action class. Given that the dependencies between states may extend over the entire image, we assume that the states and their edges form a fully-connected graph. Approximate inference is provided by a greedy algorithm that iteratively maximises over each state in turn.

Noting the number of superpixels in the $i$-th image in a training set as $T_i$, we have $x_i = [x_i^1,\ldots,x_i^t,\ldots,x_i^{T_i}]$, with $x_i^t$ a $D$-dimensional vector of scores; $h_i = [h_i^1,\ldots,h_i^t,\ldots,h_i^{T_i}]$, with $h_i^t \in \{1,\ldots,K\}$ a superpixel class; and $y_i \in \{0,1\}$ a binary variable for a given action class. We build one such model for each action class. At training time, the action class is supervised while states are hidden.

\subsection{Object detectors}
The first step of our processing pipeline is the decomposition of the image into superpixels. For this task, we use the efficient graph-based segmentation algorithm proposed by Felzenszwalb \textit{et al.}~\cite{felzenszwalb2004efficient} that was also adopted by~\cite{pei2013efficient}. This step achieves good over-segmentation of the image into regions of predominantly homgeneous nature, and errors in this process can be tolerated by the ensuing soft-assignment stage. Fig. 1 shows an example of superpixel segmentation.

To classify the superpixels, we have used the class set of the MSRC-21 dataset consisting of $23$ diverse classes from typical background and foreground objects~\cite{MSRCDataset}.
Similarly to \cite{pei2013efficient}, we use a combination of appearance-based and bag-of-features descriptors as the feature vector. The appearance-based descriptor has $51$ features, comprising of: 1) $40$ color features measuring mean, standard deviation, skewness and kurtosis of RGB, LAB, YCrCb color space channels and the gray image; and 2) $11$ texture features obtained from the application of an average filter and five different responses from Gaussian and Laplacian-of-Gaussian filters. The bag-of-features is obtained by first computing dense SIFT descriptors~\cite{vedaldi08vlfeat} in the superpixel region at three different scales and then encoding the descriptors into a dictionary of $400$ visual words learned by $K$-means clustering. We concatenate the appearance-based descriptor and the bag-of-features into a $451$-D vector, noted as $s_{i}^{t}$ for the $t$-th superpixel of the $i$-th image. 

Once the feature vector is extracted, a superpixel classifier is built by multiclass SVM~\cite{crammer2002algorithmic,TsochantaridisJMLR5} with a linear kernel trained over the MSRC-21 dataset. Please note that this dataset is a separate dataset from the action dataset and that object detectors will not be re-trained. Once the classifier is trained, for every measurement, $s_{i}^{t}$, we compute and collect the probability scores of all $23$ classes as a feature vector, $ x_i^t$:

\begin{align}
\left[ x_i^t = p(k | s_{i}^{t}) \propto \exp(w^{T}_{k} s_{i}^{t}) \right], \hspace{3pt} \forall k
\end{align}

where $w_{k}$ notes the $k$-th class' parameter vector of the SVM multiclass classifier. Such a vector of posterior probabilities will be later used as the superpixel's measurement for action classification, exploiting the semantic of the object classes and reducing the measurement dimensionality from $451$-D to $23$-D. 

\section{Latent structural SVM}

For action classification, we wish to learn the following linear prediction function:

\begin{align}
\label{eq:score}
(\bar{y},\bar{h}) = f_w(x) = \argmax_{y,h}[w^{T}\psi(x,h,y)]
\end{align}     

where $\psi(x,h,y)$ is a generalised feature function computing a combined map over measurements $x$, states $h$ and class $y$, and $w$ is a corresponding parameter vector. The action class and superpixel states are predicted jointly and typically only the predicted class, $\bar{y}$, is retained. For parameter estimation, we adopt the well-established latent structural SVM framework~\cite{yu2009learning}. This is a regularised minimum-risk framework guaranteed to provide a local optimum for structural models with latent variables. Its learning objective:

\begin{align}
\label{eq:optimisation}
\begin{split}
& w^* = \argmin_{w,\xi_{1:N}} \frac{1}{2} \left\| w \right\|^2 + C \sum_{i=1}^N \xi_i \hspace{5pt} \\
& s.t.\hspace{2pt} w^T \Psi(x_i,h_{i}^*,y_{i}) - w^T \Psi(x_i,h,y) \geq 1 - \xi_i \hspace{10pt} \\
& \hspace{4.0cm}\forall \{ y, h \} \neq \{ y_i, h_{i}^* \}
\end{split}
\end{align}

\begin{align}
\label{eq:assignment}
h_{i}^* = \argmax_{h} w^{*T} \Psi(x_i, h, y_i)
\end{align}

is an iterative objective that alternates between the constrained optimisation in (\ref{eq:optimisation}), performed using the current values for latent variables $h_{i}^*$, and a new assignment for $h_{i}^*$ (\ref{eq:assignment}) from updated model $w^{*}$.  Implementation requires a number of design choices including the definition of a suitable feature function, $\psi(x,h,y)$, the initialisation of latent variables $h$, and efficient algorithms for inference and augmented inference. These components are presented in the following sub-sections.

The learning procedure in (\ref{eq:optimisation}-\ref{eq:assignment}) can be initialised by either an arbitrary vector $w^*$ in (\ref{eq:assignment}) or an arbitrary assignment for the $h_{i}^*$ in (\ref{eq:optimisation}). Given that we have trained a multiclass superpixel classifier to obtain the feature set, the most natural choice is to initialise the states with the prediction from this classifier:
  
\begin{align}
\label{eq:initialisation}
h_{i_{init}}^{*t} = \argmax_{k} [p(k | s_i^t)]
\end{align}

The above is equivalent to initialising the states with individual predictions, delegating the discovery of correlations to the training stage. 

\subsection{Feature function and score function}
\label{subsec:score}

The features in feature function $\psi(x,h,y)$ reflect the topology of the graphical model that includes an edge between each superpixel's measurement and its state variable, a fully-connected graph amongst states, and an edge between each state and the action class. In detail, $\psi(x,h,y)$ breaks into:

\begin{itemize}
\item \textit{measurement features}, $\varphi(x^{t},h^{t}=j)$: these features map the measurement vector of the $t$-th superpixel, $x^t$ (dimensionality:  $K$) to its state, $h^{t}$ 
(possible values: $\{1 \ldots K\}$). The size of this feature vector is $K^2$ and, given $h^{t}=j$, it consists of $x^t$ starting at index $K(j - 1)$ and zero-padding elsewhere;
\vspace{0.2cm}
\item \textit{state features}, $\theta(h^{t} = j,h^{u}=k)$: these features report the co-occurrence of states $h^{t} = j$ and $h^{u} = k$. The size of this feature vector is again $K^2$, with a value $1$ at index $K(j - 1) + k - 1$ and zeros elsewhere;
\vspace{0.2cm}
\item \textit{class features}, $\phi(y = b \in \{0,1\},h^{t} = j$): these features report the co-occurrence of action class $y = b$ and state $h^{t} = j$. The size of this feature vector is $2K$, with a value $1$ at index $bK + j - 1$ and zeros elsewhere.
\end{itemize}

We refer the reader to~\cite{TsochantaridisJMLR5} for further details on feature maps. Given such feature vectors, the score function in (\ref{eq:score}) is computed as:

\begin{equation} \label{eq:scoringfunction}
\begin{split}
& w^{T}\psi(x,h,y) = \sum_{t=1}^{T} w_{\varphi}^{T} \hspace{1.5pt} \varphi(x^t,h^t)
  + \sum_{t=1}^{T} \sum_{ \substack{u=1,\\u\ne t} }^{T} w_{\theta}^{T} \hspace{1.5pt} \theta(h^{t},h^{u})\\ 
& \hspace{4.8cm} + \sum_{t=1}^{T} w_{\phi}^{T} \hspace{1.5pt} \phi(y,h^{t})
\end{split}
\end{equation}

where $w^T = \left[ w_{\varphi}^{T} \hspace{1.5pt} w_{\theta}^{T} \hspace{1.5pt} w_{\phi}^{T} \right]$ is the concatenation of the parameter vectors for the corresponding features.  


\subsection{Loss-augmented inference}

Following the structured learning approach of \cite{yu2009learning}, a fundamental step in the learning procedure is the computation of a loss-augmented version of the inference. As loss function, we simply use the 0-1 loss function:\\

\[
 \bigtriangleup(y^{gt}, y) =
  \begin{cases}
   0 & \text{if } y  \neq y^{gt} \\

   1       & \text{if } y = y^{gt}
  \end{cases}
\]

where $y^{gt}$ represents the ground-truth label. With this choice, it can easily be seen that the loss-augmented inference:

\begin{align}
  (\bar{y},\bar{h}) = \argmax_{y,h} \left[ w^{T} \psi(x,h,y) + \bigtriangleup(y^{gt},y) \right]
\end{align}

is equivalent to the standard inference in (\ref{eq:score}) with the addition of a unit score over the incorrect class. 

\section{Experimental results}
\label{exp}

We have evaluated the proposed approach on the most challenging static action recognition dataset released to date, Stanford 40 Actions~\cite{yao2011human}. This dataset contains images of humans performing 40 different classes of actions, including visually-challenging cases such as ``fixing a bike'' versus ``riding a bike'' or ``phoning'' versus ``texting message''; the full class list is provided in the annotation of Fig.~2. The number of samples per class varies between $180$ and $300$, for a total of $9,532$ images. A standard training/test split is made available by the authors on their website, selecting 100 images from each class for training and leaving the remaining for testing.\\

Training single-class classifiers, also referred to as detectors, using all the available training samples leads to a very unbalanced set over the positive and negative classes ($100$ and $3,900$ samples, respectively). In the case of a conventional SVM objective as in (\ref{eq:optimisation}), this biases the prediction function towards negative predictions. For this reason and in order to save learning time, we decided to sub-sample the negative training samples of each classifier by randomly choosing $5$ images from each of its $39$ negative classes. Parameter $C$ in (\ref{eq:optimisation}) was set to $1$.\\

When measuring the performance of a detector, it would be trivial to achieve high overall accuracies by always predicting the negative class. Therefore, a dataset like Stanford 40 Actions requires measuring the accuracy for the positive and negative classes separately, or providing measures such as precision and recall, average precision at various levels of recall, or similar. In this work, we decided to report the accuracy for the positive and negative classes as the main figure, as shown in Fig.~2. 
The mean accuracy over all the $40$ classes proved $74.06\%$ for the positive class and $88.50\%$ for the negative class. These results prove that the individual detectors are well balanced over positive and negative predictions and show that the overall accuracy is much higher, for instance, than that recently reported in~\cite{IkizlerECCVW2012} ($55.93\%$). For the individual classes, some classes reach very high accuracy over both positive and negative samples. For instance, class ``rowing a boat'' achieves $93.90\%$ and $96.15\%$ accuracy, respectively. Other classes show significant missed detections: for instance, class ``running'' achieves only $48.99\%$ accuracy over the positive samples, arguably as it is hard to recognise a running action from a static frame. Overall, these results seem even more remarkable considering that the $23$ object detectors were trained on a completely separate dataset (no re-training was performed on the action dataset) and from classes mostly unrelated with the objects portrayed in Stanford 40 Actions (such as bikes, phones, cameras, telescopes, microscopes and several others). 


\begin{figure}[t]
\label{fig:results}
\begin{minipage}[b]{1.0\linewidth}
\centering
\centerline{\label{a}\includegraphics[width=9.5cm,height=9.0cm]{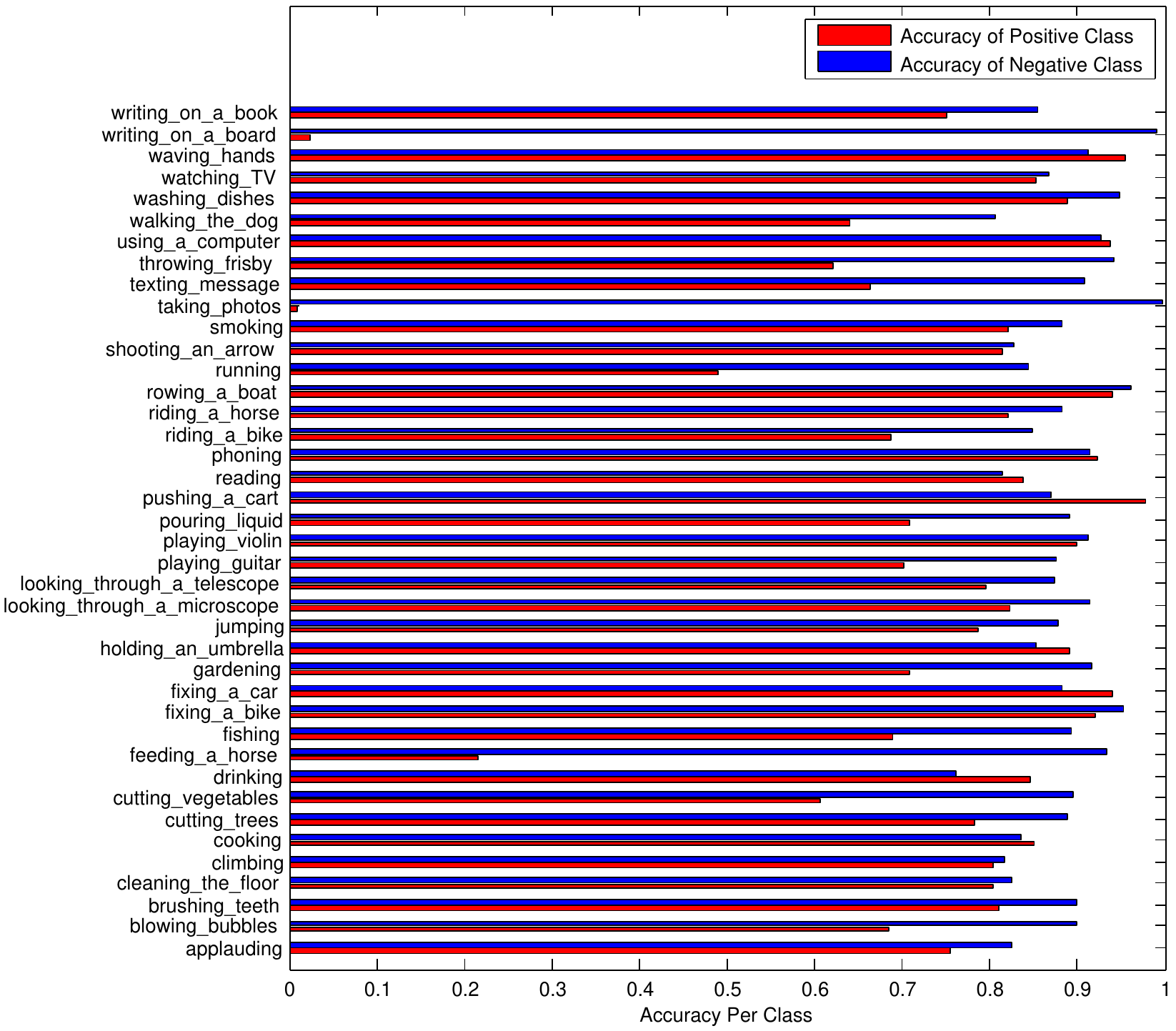}}
\caption{Achieved accuracy over the positive and negative class for all classes in the Stanford 40 Actions dataset.}  
\end{minipage}
\end{figure}

\section{Conclusion}

In this paper, we have proposed an approach to action recognition in still images leveraging an intermediate superpixel representation of the image. The approach consists of two stages: in a first stage, the image is segmented into a set of superpixels and an array of trained object detectors is applied to each superpixel to extract a vector of detector scores. In a second stage, the score vectors are used as measurements in a graphical model that jointly predicts the superpixels' classes together with the action class. Experiments conducted over the highly challenging Stanford 40 Actions dataset have resulted in a remarkable accuracy of $74.06\%$ for the positive class and $88.50\%$ for the negative class averaged over the $40$ action classifiers. These results give evidence to the existence of a useful relationship between the classes of the image superpixels and that of the main action. Possible ways to further improve the performance of the proposed model would be to adopt a larger, more universal set of object detectors, or a set of detectors tuned in the specific object classes of given action datasets.



\bibliographystyle{IEEEbib}
\bibliography{References}

\begin{thebibliography}{10}

\bibitem{LaptevBMVC2010}
V.~Delaitre, I.~Laptev, and J.~Sivic,
\newblock ``Recognizing human actions in still images: a study of
  bag-of-features and part-based representations,''
\newblock in {\em Proceedings of the British Machine Vision Conference}, 2010,
  pp. 1--11.

\bibitem{IkizlerICPR2008}
Nazli Ikizler, R.~Gokberk Cinbis, Selen Pehlivan, and Pinar Duygulu,
\newblock ``Recognizing actions from still images,''
\newblock in {\em Proceedings of the 19th International Conference on Pattern
  Recognition, 2008, ICPR 2008}, 2008, pp. 1--4.

\bibitem{laptev2005space}
I.~Laptev,
\newblock ``{On space-time interest points},''
\newblock {\em International Journal of Computer Vision}, vol. 64, no. 2, pp.
  107--123, 2005.

\bibitem{Dalal2005}
Navneet Dalal and Bill Triggs,
\newblock ``Histograms of oriented gradients for human detection,''
\newblock in {\em International Conference on Computer Vision \& Pattern
  Recognition}, June 2005, vol.~2, pp. 886--893.

\bibitem{lowe2004distinctive}
D.G. Lowe,
\newblock ``{Distinctive image features from scale-invariant keypoints},''
\newblock {\em International journal of computer vision}, vol. 60, no. 2, pp.
  91--110, 2004.

\bibitem{Oliva2001}
Aude Oliva and Antonio Torralba,
\newblock ``Modeling the shape of the scene: A holistic representation of the
  spatial envelope,''
\newblock {\em Int. J. Comput. Vision}, vol. 42, no. 3, pp. 145--175, May 2001.

\bibitem{IkizlerICCV2009}
N.~Ikizler-Cinbis, R.G. Cinbis, and S.~Sclaroff,
\newblock ``Learning actions from the web,''
\newblock in {\em 2009 IEEE 12th International Conference on Computer Vision}.
  IEEE, 2009, pp. 995--1002.

\bibitem{GuptaPAMI2009}
A.~Gupta, A.~Kembhavi, and L.S. Davis,
\newblock ``Observing human-object interactions: Using spatial and functional
  compatibility for recognition,''
\newblock {\em IEEE Transactions on Pattern Analysis and Machine Intelligence},
  vol. 31, no. 10, pp. 1775 --1789, oct. 2009.

\bibitem{LiFeiFeiCVPR2010}
Bangpeng Yao and Li~Fei-Fei,
\newblock ``Grouplet: A structured image representation for recognizing human
  and object interactions,''
\newblock in {\em IEEE Conference on Computer Vision and Pattern Recognition,
  CVPR 2010}, 2010, pp. 9--16.

\bibitem{MoriCVPR2010}
Weilong Yang, Yang Wang, and Greg Mori,
\newblock ``Recognizing human actions from still images with latent poses,''
\newblock in {\em CVPR}, 2010, pp. 2030--2037.

\bibitem{LiFeiFeiICCV2011}
Bangpeng Yao, Xiaoye Jiang, Aditya Khosla, Andy~Lai Lin, Leonidas~J. Guibas,
  and Li~Fei-Fei,
\newblock ``Action recognition by bases of action attributes and parts,''
\newblock in {\em International Conference on Computer Vision (ICCV)}, 2011,
  pp. 1331--1338.

\bibitem{PictorialStructures1973}
M.~Fischler and R.~Elschlager,
\newblock ``The representation and matching of pictorial structures,''
\newblock {\em IEEE Transactions on Computers}, vol. 22, no. 1, pp. 67--92,
  Jan. 1973.

\bibitem{PedroPAMI2010}
P.~F. Felzenszwalb, R.~B. Girshick, D.~McAllester, and D.~Ramanan,
\newblock ``Object detection with discriminatively trained part-based models,''
\newblock {\em IEEE Transactions on Pattern Analysis and Machine Intelligence},
  vol. 32, no. 9, pp. 1627--1645, Sept. 2010.

\bibitem{Guo2014}
Guodong Guo and Alice Lai,
\newblock ``A survey on still image based human action recognition,''
\newblock {\em Pattern Recognition}, vol. in press, no. 0, pp. --, 2014.

\bibitem{felzenszwalb2004efficient}
Pedro~F Felzenszwalb and Daniel~P Huttenlocher,
\newblock ``Efficient graph-based image segmentation,''
\newblock {\em International Journal of Computer Vision}, vol. 59, no. 2, pp.
  167--181, 2004.

\bibitem{pei2013efficient}
Deli Pei, Zhenguo Li, Rongrong Ji, and Fuchun Sun,
\newblock ``Efficient semantic image segmentation with multi-class ranking
  prior,''
\newblock {\em Computer Vision and Image Understanding}, 2013.

\bibitem{MSRCDataset}
A~Criminisi,
\newblock ``Microsoft research cambridge object recognition image
  database,http://research.microsoft.com/en-us/projects/objectclassrecognition/,''
  2004.

\bibitem{vedaldi08vlfeat}
A.~Vedaldi and B.~Fulkerson,
\newblock ``{VLFeat}: An open and portable library of computer vision
  algorithms,'' \url{http://www.vlfeat.org/}, 2008.

\bibitem{crammer2002algorithmic}
Koby Crammer and Yoram Singer,
\newblock ``On the algorithmic implementation of multiclass kernel-based vector
  machines,''
\newblock {\em The Journal of Machine Learning Research}, vol. 2, pp. 265--292,
  2002.

\bibitem{TsochantaridisJMLR5}
I.~Tsochantaridis, T.~Joachims, T.~Hofmann, and Y.~Altun,
\newblock ``Large margin methods for structured and interdependent output
  variables,''
\newblock {\em JMLR}, vol. 6, pp. 1453--1484, 2005.

\bibitem{yu2009learning}
Chun-Nam~John Yu and Thorsten Joachims,
\newblock ``Learning structural svms with latent variables,''
\newblock in {\em Proceedings of the 26th Annual International Conference on
  Machine Learning}. ACM, 2009, pp. 1169--1176.

\bibitem{yao2011human}
Bangpeng Yao, Xiaoye Jiang, Aditya Khosla, Andy~Lai Lin, Leonidas Guibas, and
  Li~Fei-Fei,
\newblock ``Human action recognition by learning bases of action attributes and
  parts,''
\newblock in {\em Computer Vision (ICCV), 2011 IEEE International Conference
  on}. IEEE, 2011, pp. 1331--1338.

\bibitem{IkizlerECCVW2012}
Fadime Sener, Cagdas Bas, and Nazli Ikizler-Cinbis,
\newblock ``On recognizing actions in still images via multiple features,''
\newblock in {\em Computer Vision – ECCV 2012. Workshops and Demonstrations},
  2012, vol. 7585 of {\em Lecture Notes in Computer Science}, pp. 263--272.

\end{thebibliography}

\end{document}